%% file: 00_root.tex
\documentclass[conference]{IEEEtran}
\IEEEoverridecommandlockouts
\usepackage{cite}
\usepackage{amsmath,amssymb,amsfonts}
\usepackage{algorithmic}
\usepackage{graphicx}
\usepackage{textcomp}
\usepackage[table]{xcolor}
\usepackage{gensymb}
\usepackage{array} 
\usepackage{soul}
\begin{document}

\title{A deep reinforcement learning based motion cueing algorithm for vehicle driving simulation\\
}

\author{\IEEEauthorblockN{Hendrik Scheidel}
	\IEEEauthorblockA{\textit{Institute of System Dynamics and Control, Space Systems Dynamics,}\\
		\textit{Deutsches Zentrum für Luft- und Raumfahrt (DLR)}\\
		\textit{German Aerospace Center}\\
		82234, Germany\\
		hendrik.scheidel@dlr.de}
}

\author{
	Hendrik Scheidel,
	Houshyar Asadi, \IEEEmembership{Member, IEEE},
	Tobias Bellmann, 
	Andreas Seefried,
	Shady Mohamed, \\
	Saeid Nahavandi, \IEEEmembership{Fellow, IEEE}
	\thanks{Copyright (c) 2015 IEEE. Personal use of this material is permitted. However, permission to use this material for any other purposes must be obtained from the IEEE by sending a request to pubs-permissions@ieee.org. \linebreak}
	\thanks{H. Scheidel, H. Asadi, S. Mohamed and S. Nahavandi are with the Institute
		for Intelligent Systems Research and Innovation, Deakin University,
		Geelong, VIC 3216, Australia 
		(e-mail: hscheidel@deakin.edu.au; houshyar.asadi@deakin.edu.au; shady.mohamed@deakin.edu.au; saeid.nahavandi@deakin.edu.au).}
	\thanks{H. Scheidel, T. Bellmann, and A. Seefried are with the Institute of System Dynamics and Control, German Aerospace Center, 
		82234 Weßling, Germany 
		(e-mail: hendrik.scheidel@dlr.de; tobias.bellmann@dlr.de; andreas.seefried@dlr.de).}
}

\maketitle

\begin{abstract}
Motion cueing algorithms (MCA) are used to control the movement of motion simulation platforms (MSP) to reproduce the motion perception of a real vehicle driver as accurately as possible without exceeding the limits of the workspace of the MSP. Existing approaches either produce non-optimal results due to filtering, linearization, or simplifications, or the computational time required exceeds the real-time requirements of a closed-loop application. This work presents a new solution to the motion cueing problem, where instead of a human designer specifying the principles of the MCA, an artificial intelligence (AI) learns the optimal motion by trial and error in interaction with the MSP. To achieve this, a well-established deep reinforcement learning (RL) algorithm is applied, where an agent interacts with an environment, allowing him to directly control a simulated MSP to obtain feedback on its performance. The RL algorithm used is proximal policy optimization (PPO), where the value function and the policy corresponding to the control strategy are both learned and mapped in artificial neural networks (ANN). This approach is implemented in Python and the functionality is demonstrated by the practical example of pre-recorded lateral maneuvers. The subsequent validation shows that the RL algorithm is able to learn the control strategy and improve the quality of the immersion compared to an established method. Thereby, the perceived motion signals determined by a model of the vestibular system are more accurately reproduced, and the resources of the MSP are used more economically.
\end{abstract}


\input{01_Introduction}
\input{02_Methodology}
\input{03_Results_and_Discussion}
\input{04_Conclusion}

\bibliography{../../01_General/00_Bib_Files/Paper}
\end{document}

%% file: 01_Introduction.tex
\section{Introduction}

Motion simulation platforms (MSPs) are common tool in many industries such as aerospace or automotive because of their cost-effective and safe application. They are gaining importance in a variety of different fields including scientific research, industrial development, entertainment or the training of drivers and pilots. Compared to non-moving simulators, the additional reproduction of motion cues leads to a significant increase in the quality of the overall simulation, as shown in several studies \cite{1985Buckingham_CONF, 2000Boer, 2008Fischer}.

\begin{figure}
	\includegraphics[width=\columnwidth]{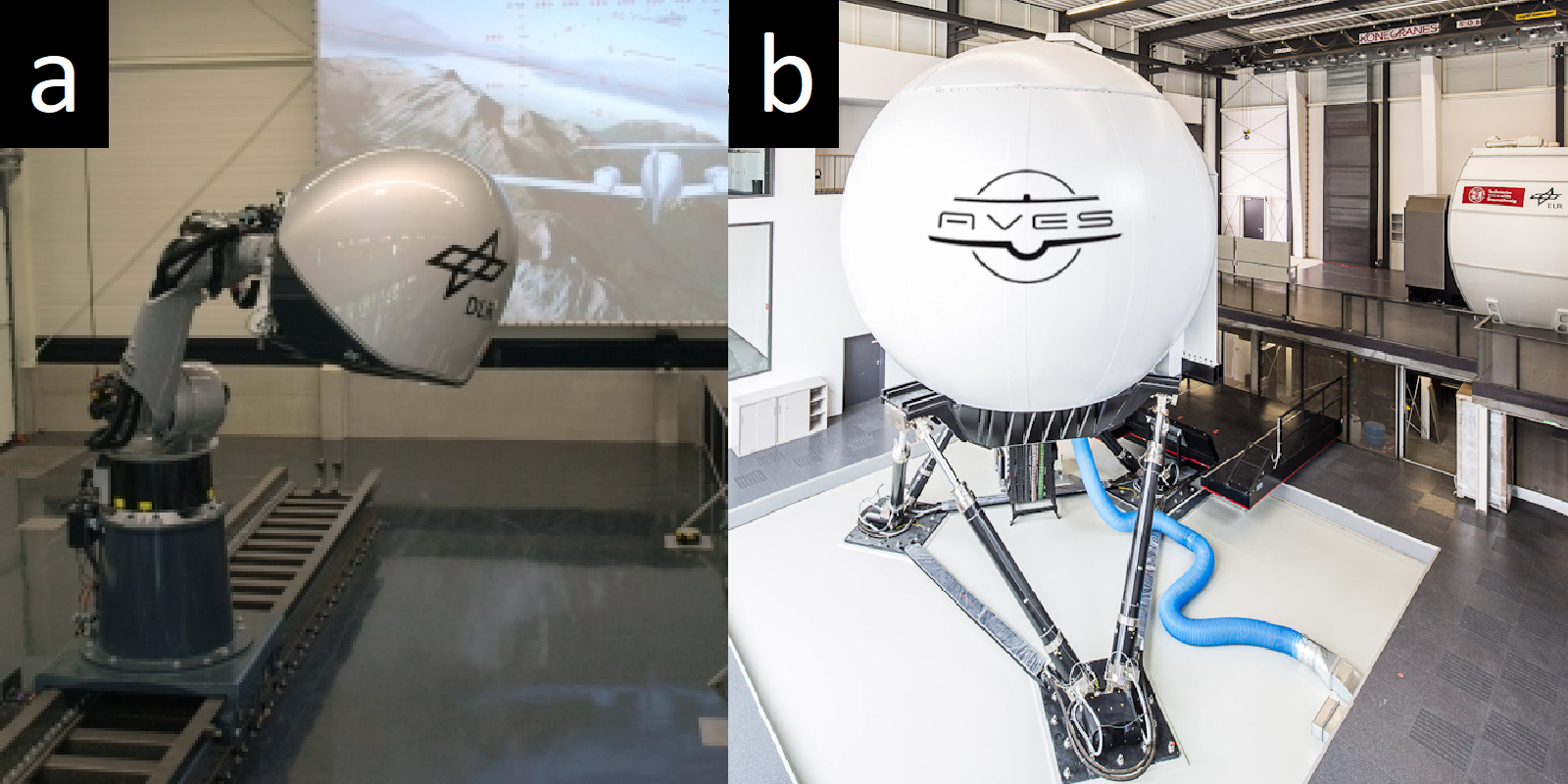}
	\caption{The \textit{Robotic Motion Simulator} at DLR Institute of System Dynamics and Control (a)~\cite{2011Bellmann_CONF} and the \textit{AVES Research Flight Simulator} at DLR Institute of Flight Systems (b)~\cite{2013Duda}.} 
	\label{RMS_vs_Stewart}
\end{figure}

Fig.~\ref{RMS_vs_Stewart} shows two common arrangements of MSPs capable of performing movements in all six degrees of freedom (DOF). In Fig.~\ref{RMS_vs_Stewart}a a serial arrangement of the actuators based on the application of a robotic arm is shown. In comparison to the more traditional hexapod arrangement, as displayed in Fig.~\ref{RMS_vs_Stewart}b, this robotic motion platform (RMP) provides a larger workspace at a very competitive price, but results in a reduced maximal payload due to its serial configuration. Furthermore, in order to utilize the available workspace, more complex control strategies are required and they have limited capability in producing high frequency motions and vibrations compared with hexapods \cite{2011Bellmann_CONF, 2019Silva}. In addition to these two variants, a variety of less common implementations and different combinations of systems exists \cite{2015Mohajer, 2016Miermeister}. With the rapid development of technologies such as VR head-mounted displays and the ongoing process of digitization, smaller MSPs with reduced DOF are also getting more popular for commercial and non-commercial applications. 

To achieve a clearer definition of the challenging task of reproducing motion cues, an understanding of human motion perception is required. The main human body part in charge of perceiving motion is located in the inner ear, known as vestibular system. It consists of semicircular canals and otoliths with the former sensing linear accelerations and the latter sensing angular velocities \cite{2016Asadi_ARTCa, 2017Asadi_ARTC}. Besides the vestibular system, further decisive element for the spatial orientation and motion perception of human are the visual and the auditory system. If there are major contradictions in the perception of these systems, disorientation and the phenomenon of motion sickness may occur, which can result in nausea, dizziness, headache and fatigue \cite{2019Iskander_ARTC}. 

A crucial role in minimizing the discrepancy of the signals is played by the control strategy used to define the MSP's movements in the limited workspace. These strategies are commonly called motion cueing algorithms~(MCAs). The classical washout (CW), representing the origin of MCAs \cite{1970Schmidt,1985Reid_ARTC,1986Reid_ARTC}, is based on the idea of limiting, scaling and filtering the signals. By using high-pass filters, the high-frequency components of the signals can be reproduced by translational and rotational motion of the simulator. One of the characteristics of the filters is a constant washout of the signals, which causes the simulator to be continuously returning to the initial position. To reproduce the low-frequency components of the translational acceleration, a rotation of the cockpit is performed in accordance with the principle of the somatogravic illusion \cite{2006MacNeilage_ARTC}. This process is generally referred to as tilt coordination. The advantages of the CW lie in its simplicity and safety, the clear definition of the physical meaning of the parameters and the capability of real-time application. On the negative side, the performance of the CW depends heavily on the experience of the system engineer. In the tuning process, the parameters of the filters are adjusted to the worst-case scenario, resulting also in inefficient use of the workspace \cite{1997Grant_ARTC}. Approaches to solve these problems are provided by the concepts of 
optimal \cite{1982Sivan_ARTC,2016Asadi_ARTC, 2022Qazani_ARTC} and adaptive control \cite{1975Parrish_ARTC, 2021Qazani_ARTCa, 2022Asadi_ARTC}.
Applying the concept of model predictive control (MPC) to the field of motion simulation leads to a new category of MCAs \cite{2004Dagdelen_CONF,2009Dagdelen_ARTC, 2021Seefried_DISS}. These algorithms use the current states of the system to predict its future behavior based on a system model in order to derive the best motion from it in turn. Therefore, an optimizer is applied to minimize a cost function which quantifies the quality of the motion by considering a mathematical model of the vestibular system. For most implementations, this process is computational highly demanding, which means that real-time capability is a major challenge.

Due to the aforementioned disadvantages of existing MCAs, especially complex workspaces that are insufficiently described by a cube approximation cannot be optimally exploited in real-time applications. This is the case when using an RMP with its highly non-linear workspace. Therefore, the application of artificial neural networks (ANNs) and reinforcement learning (RL) to address the problem of an MCA is proposed.
The application of ANNs in the field of MCAs opens up further possible solutions. Mohammadi et al. \cite{2016Mohammadi} proposed an ANN-based future reference trajectory prediction to optimize the performance of an MPC algorithm. Qazani et al. \cite{2021Qazani_ARTC,2021Qazani_CONF} further elaborated on this idea by comparing the implementation of a time-delay feedforward NN, a recurrent NN (RNN), a nonlinear autoregressive model and a type-2 quantum fuzzy NN for predicting future driving scenarios. Qazani et al.~\cite{2020Qazani_ARTC} used future driving scenario prediction by an ANN combined with fuzzy logic to preposition the platform center. Rengifo et al.~\cite{2018Rengifo_CONF} employ an RNN as a more time efficient quadratic problem solver to minimize the cost function of an MPC algorithm.
In the work of Koyuncu et al. \cite{2020Koyuncu_CONF}, an ANN is trained to imitate the behavior of an open-loop MPC algorithm. For this purpose, the MPC algorithm with unlimited prediction horizon is used to determine the optimal simulator position and orientation specifications for prerecorded driving trajectories by minimizing a cost function. This generated data is then used to train an ANN via supervised learning. It can be shown that the trained ANN can be used as an MCA on both the training data and on arbitrary trajectories during driver-in-the-loop application. However, the separation of the optimization by MPC with unlimited horizon and the subsequent approximation by ANN provides an additional step that can be circumvented by applying the concept of deep RL. Besides the general differences of deep RL and MPC \cite{2009Ernst_ARTC, 2021Lin_ARTC} the additional approximation errors can be avoided which may finally lead to an increase of the quality of the motion simulation.

In the present work, the first fully ANN-based MCA realized through deep RL is proposed. The use of deep RL combines the advantages of ANNs as a universal nonlinear function approximator, which enables arbitrary control strategies, with the advanced computational approach of automated goal-directed problem solving and decision making through interaction learning. During the training process, the RL agent interacts directly with the MSP, learning both the control strategy and the behavior of the system in terms of a value function. The control strategy, referred to as a policy, corresponds to an fully ANN-based MCA. The solution found in the process is therefore not explicitly specified by a human engineer but found autonomously by interacting with the system and evaluating the given feedback. This demonstrates a core ability of RL to independently make creative decisions that are not based on human knowledge and choices \cite{2016Silver_ARTC}. 

The use of the ANN-based MCA obtained by deep RL solves several problems of established algorithms. Compared to filter based MCAs there is an unlimited control variety and inherent disadvantages of filtering like the phase shift caused by the low pass filter can be avoided. The execution of the algorithm corresponding to the evaluation of the trained ANN is performed by using a small amount of computational resources, so that the real-time requirements can be easily satisfied. This is a major challenge for software and hardware in MPC-based MCAs. In addition, the intelligent and creative behavior attributed to artificial intelligence could lead to a better understanding of complex workspaces. Especially if the motion simulation is realized by an RMP, this may result in a more efficient usage of the workspace. 

Outside the field of motion simulation, RL has already been successfully applied in many areas. Among these, the general application of RL to robotic serial manipulators represents a frequently contemplated application. Typically, these are manipulation or positioning tasks where the robot is trained to pick up an object or move it to a desired position \cite{2013Kober_ARTC}. Another common use case is the implementation of RL for path-following control of autonomous vehicles such as in~\cite{2019Ultsch, 2020Ultsch}. To the best of the authors' knowledge, however, there is at present no application of RL algorithms to acceleration trajectories in robotic field or specifically in the area of motion simulation.

In section II, the proposed method and approaches are described. Section III shows the training of the new interactive RL-based MCA for a specific application and presents the results of a validation. Finally, section IV provides the conclusion of the performed research study.

%% file: 02_Methodology.tex
\section{Methodology}
First, the theoretical background of the applied RL algorithm is presented. Then the designed framework required for the process of training and the process itself are described.

\subsection{Reinforcement Learning}
RL gained a lot of attention in the recent years. In 2017, Silver et al.'s AlphaGo program, based on supervised and reinforcement learning, achieved great popularity \cite{2016Silver_ARTC, 2017Silver_ARTC}. It was the first computer program that managed to win in the highly complex game of Go against a multiple world champion. Along with robotics, computer and board games are one of the main applications of RL \cite{2013Mnih_ARTC}. However, the great potential of RL has already been demonstrated in many other areas, such as health-care, finance or plant management. The term deep RL refers to use of deep ANNs as function approximators. In the following, the theory of the deep RL algorithms applied in this work will be discussed. Most of it is taken from the groundwork by Sutton and Barton \cite{2018Sutton_BOOK}.

The concept of RL is based on the theoretical mathematical formulation of a Markov decision process (MDP). This is used to model a decision problem where an agent interacts with an environment via a defined interface. In a time-discrete sequence the agent receives the state vector $S_t \in \mathcal{S}$ of the environment and reacts with an action vector $A_t \in \mathcal{A}$, through which it is able to interact with the environment. Here $\mathcal{S}$ defines the set of possible states and $\mathcal{A}$ the set of possible actions, respectively.
In the following time step, the agent receives the resulting new state $S_{t+1}$ of the environment and a numerical reward $R_{t+1}$, that evaluates the quality of the state. Based on the new state $S_{t+1}$, the agent then chooses the new action $A_{t+1}$. By continuing in this alternating pattern, this process leads to a sequence or trajectory of $\tau=(S_{0}, A_{0}, S_{1}, R_{1}, A_{1}, S_{2}, R_{2}, A_{2},...)$ . The dynamics of an MDP is described by the transition probability $T(S_{t+1}|S_{t},A_{t})$. This specifies the probability, unknown to the agent, of the stochastic environment to transit from the current state $S_{t}$ to the next state $S_{t+1}$ depending on the action $A_{t}$.

During the training process, the agent explores the behavior of the environment through interaction and thus independently collects its training data. Thereby, it tries to gain experience about the behavior of the environment in order to develop a control strategy $\pi (A_t|S_t)$, referred to as policy. This policy, corresponding to a mapping of states to actions, can be represented in the form of an ANN. In this case, the notation $\pi_\theta$ is usually used, where $\theta$ denotes to the parameters (weights and biases) of the ANN. The dynamics of the overall system thus depends on the shared probability distribution $p_\theta$ composed of the policy and the transition probability:

\begin{equation}
	p_\theta (\tau) = \mu_0(S_0) \prod_{t=1}^\infty \pi_\theta (A_t | S_t) \; T(S_{t+1}|S_t, A_t),
\end{equation}

where $\mu_0$ corresponds to the start probability distribution of the first state of the environment. The overall goal of RL can be formalized as finding the optimal policy $\pi^*$, that is receiving the highest cumulative discounted reward $G(\tau)$ for all possible trajectories:

\begin{equation}
	G(\tau) = \sum_{t=0}^\infty \gamma^t R_{t+1},
\end{equation}

where $\gamma$ is the so called discount factor, $0 \le \gamma \le 1$. It determines the sensitivity of the agent for long term rewards.
RL provides several approaches and algorithms to achieve this goal. In this work proximal policy optimization (PPO)~\cite{2017Schulman_ARTC}, a policy gradient method based on an actor-critic implementation, is used. The concept and the mathematical foundation of this algorithm is explained in the following. 

Training data is obtained episodically by stochastic variation of the current policy. Based on the information contained, the policy is adjusted in the direction of the gradient of an objective function and new training data is collected using the updated policy. The mathematical representation of the idea is centered on classical policy gradient methods~\cite{1992Williams}. Here, the objective function $L(\theta)$ is defined as the expected value, $\mathbb{E}$, of the discounted reward $G(\tau)$ for a trajectory $\tau$ sampled from the probability distribution $p_\theta$. It can be formulated as:

\begin{equation}
	L(\theta) = \mathbb{E}_{\tau \sim p_\theta (\tau)} \Bigg[ \sum_{t=0}^\infty \gamma^t R_{t+1}\Bigg]
	= \mathbb{E}_{\tau \sim p_\theta (\tau)} [G(\tau)].
\end{equation}

The further derivation of the algorithm is based on the calculation of the gradient of $L(\theta)$ in dependence of the parameters $\theta$. For a detailed derivation, the reader is referred to the basic works by Sutton and Barto~\cite{2018Sutton_BOOK} and Dong et al.~\cite{2020Dong}. The gradient can be formulated as:

\begin{equation}
	\nabla_\theta L (\theta) = \mathbb{E}_{\tau \sim p_\theta (\tau)}
	\Bigg[
	G(\tau) \sum_{t=0}^\infty \nabla_\theta \text{log} [\pi_\theta(A_t|S_t)]
	\Bigg].
\end{equation}

In the actor-critic implementation of Stable-Baselines3~\cite{2021Raffin_ARTC} which is used here, the collected training data is exploited to train a second ANN in addition to training the policy function. Its task is to approximate the expected value of a state, the value function $V_\zeta(S_t)$, acting in the role of the critic, defined by the parameters $\zeta$. This additional information can then be used as a baseline to determine the advantage of received rewards over the expected value and thus to optimize the gradient of the objective function. Mathematically, this corresponds to reducing the variance of the data.

What differentiates PPO, the algorithm used here, from other policy-gradient methods is the way of controlling the margin of the update performed on the policy throughout the training. This has a decisive influence on the progress and success of a training. If the step size is too small, convergence problems can occur; if it is too large, the training can become unstable. PPO solves this problem by limiting $\xi(\theta) = \frac{\pi_\theta}{\pi_{\theta,\text{old}}}$, the ratio of the new to the old policy. If the value of the ratio is $1$, the policy has not been changed. The parameter $\eta$ defines how far $\xi$ can deviate form the value $1$ (i.e. the size of the maximum policy update). A typical value is $0.2$~\cite{2017Schulman_ARTC}. The ratio of policies $\xi(\theta)$ is thus clipped to values between $1 - \eta = 0.8$ and $1 + \eta = 1.2$.

\subsection{RL-Based MCA}
The formulation of the framework and the process of training is mainly based on the definition of the MDP. This defines the behavior of the environment and the interfaces between it and the agent. Like the MSP, the framework itself is not limited to any specific vehicle class. Therefore, the person in charge of controlling the vehicle is referred to as a pilot, regardless of the specific vehicle class. Since in the process of training, both, information about the state of the MSP and about the target trajectory from the simulated vehicle are used, a clear notation is needed. Information concerning the MSP is given the notation~'m' and information of the target trajectory from the simulated vehicle is given the notation~'v'.

\begin{figure}
	\includegraphics[width=\columnwidth]{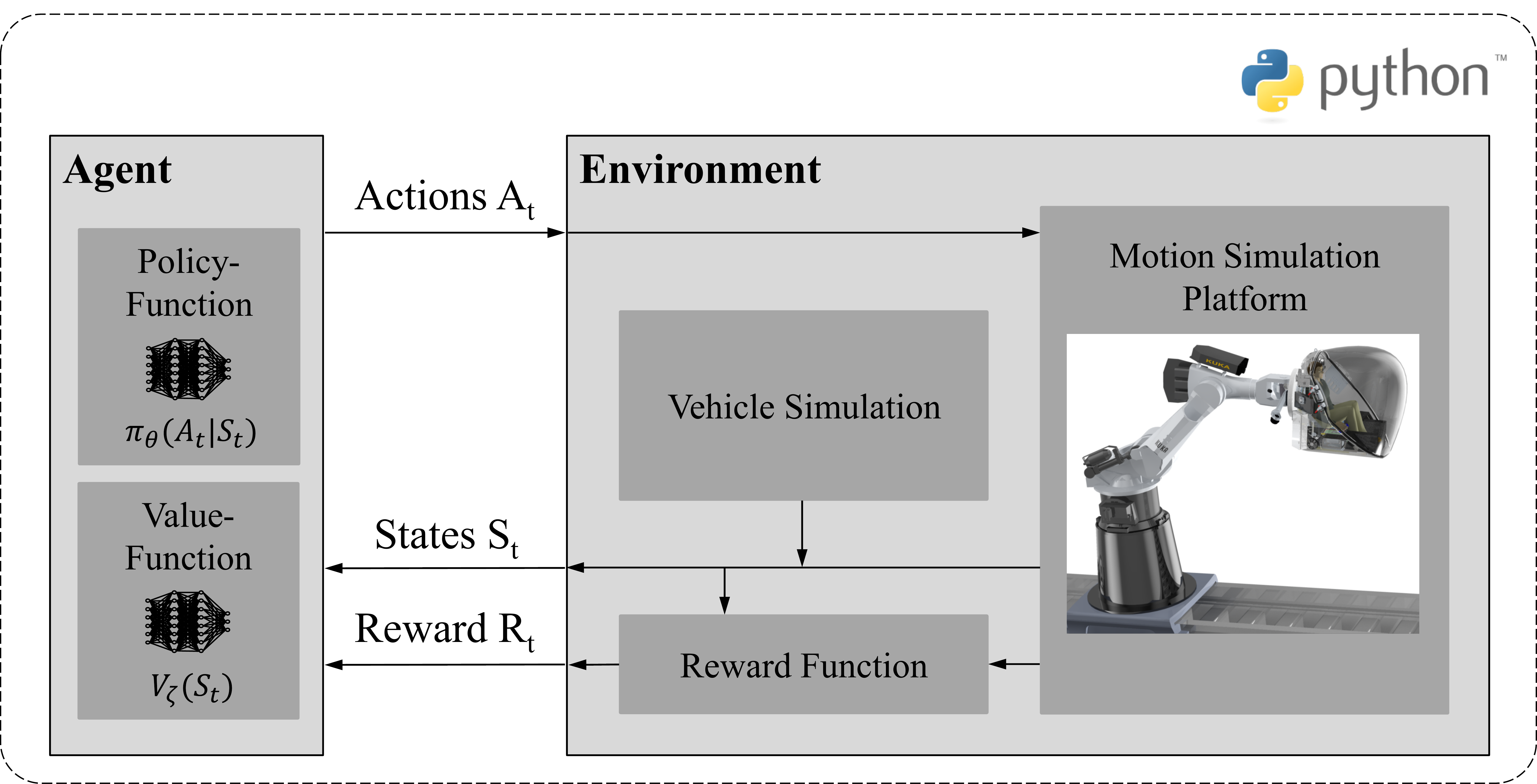}
	\caption{Block Diagram showing the structure of the MDP. At each time step, the agent first defines an action $A_t$ and then receives the resulting state $S_t$ and reward $R_t$.}
	\label{Block-Diagram}
\end{figure}

As shown in Fig.~\ref{Block-Diagram} the environment consists of the MSP, the Vehicle Simulation and the Reward Function. The agent controls the MSP in discrete time steps and the resulting motion is compared to a variable reference trajectory provided by the vehicle simulation. The resulting MDP and the interface with the agent are discussed in the following.
The agent defines the action vector $A_t$ at each time step. This information is interpreted and processed in the environment. Here it is used to specify the motion of a simulative representation of an MSP. Direct definition of the position $x_t^\text{m}$ and the orientation $\varphi_t^\text{m}$ of the pilot with in the action vector has shown not to be suitable in combination with the stochastic exploration in the RL training process. For example, a random definition of the position at each time step leads to extreme, non-feasible, accelerations which in turn significantly complicates the interpretability of the states and the reward and thus leads to a more unstable training process. Therefore, the rate of change of the main objectives $a_t^\text{m}$ and $\omega_t^\text{m}$ are specified.
The action vector $A_t$ is thus defined as:

\begin{equation}
	A_t = [\Delta a_t^\text{m},\Delta \omega_t^\text{m}]^T.
\end{equation}

These actions are integrated and processed by a simulated representation of the MSP which in return provides the translational acceleration $a_t^\text{m}$, velocity $v_t^\text{m}$, position $x_t^\text{m}$ and the angular velocity $\omega_t^\text{m}$ and orientation $\varphi_t^\text{m}$ of the simulator, which are applied on the pilot. In addition to this, the superposition resulting from the combination of gravitation and acceleration, the so-called specific force $f_t^\text{m} = a_t^\text{m} - g_t^\text{m}$, is provided. This corresponds to the total force acting on the pilot.

Furthermore, a coupled vehicle simulation provides the specific force $f_{t+1}^\text{v}$  and angular velocity $\omega_{t+1}^\text{v}$ acting on the pilot in the simulated vehicle scenario at the next time step. These values are bundled with the information of the motion simulator to a state vector $S_t$:

\begin{equation}
	\begin{split}
		S_t = [x_t^\text{m}, v_t^\text{m}, a_t^\text{m}, f_t^\text{m}, \varphi_t^\text{m}, \omega_t^\text{m}, f_{t+1}^\text{v}, \omega_{t+1}^\text{v}]^T.
	\end{split}
\end{equation}

After normalization of all values, they are passed to the RL agent.

Simultaneously, they are used as input to the reward function. As described in the previous section, the numerical feedback coming from the reward function constitutes the optimization criterion of the training process. It therefore serves as a mathematical formulation of the overall goal. Here, the designed reward function consists of seven components subtracted from an arbitrarily set maximum value of $1$ per time step:

\begin{equation}
	\label{eq:rl_reward}
	R_t = 1 - r_{f} - r_{\omega} - r_{F} - r_x - r_{ws,x} - r_{ws,\varphi}.
\end{equation}

Each of them is weighted by a coefficient $\text{w}_i$ to normalize and prioritize the different goals. In the following, the definition and meaning of the individual components is explained. The first two components aim to award minimization of deviation from target values. Whereby the first component $r_f$ is concentrated on the translation channel. The absolute difference of the specific forces $f_t^{r}$ and $f_t^{v}$ acting on the pilot in the motion simulator and on the driver in the vehicle is formed:

\begin{equation}
	r_f = \text{w}_f (|f_t^\text{m} - f_t^\text{v}|).
\end{equation}

Accordingly, the second component $r_\omega$ considers the absolute difference between the angular velocities applied in the motion simulator $\omega_t^\text{m}$ and in the vehicle $\omega_t^\text{v}$. In contrast to the first component, a margin of tolerance corresponding to the perception threshold around the roll axis of $\omega_\text{PT} = 3.0 \frac{\text{degree}}{s}$ \cite{1985Reid_ARTC} is permitted here to allow a certain rate of rotation to be free of penalty. This encourages the realization of the tilt coordination:

\begin{equation}
	r_\omega =
	\begin{cases}
		0 & \text{if } |\omega_t^\text{m} - \omega_t^\text{v}| \leq \omega_\text{PT} \\
		\text{w}_\omega (|\omega_t^\text{m} - \omega_t^\text{v}|) & \text{else}.
	\end{cases}
\end{equation}

Next, the directional fidelity of the specific forces  $f_t^{r}$ and $f_t^{v}$ is assessed. This takes into account that the influence of wrong or missing motion cues on motion perception is more significant than that of an error in the scaling of a cue. For this purpose, the function $F(f_t^{r}, f_t^{v})$ evaluates for the specific forces whether one of the following three cases is given: both in positive direction, both in negative direction or both $0$ within a tolerance range. If one of these cases is present, the function takes the value $0$, if this is not the case, it receives the value $1$. The contribution to the reward function is then calculated as:

\begin{equation}
	r_F = \text{w}_F F(f_t^{r}, f_t^{v}).
\end{equation}

The fourth component penalizes a deviation from the starting point $x_t^r=0.0$ to encourage a return to it:

\begin{equation}
	r_x = \text{w}_x x_t^\text{m}.
\end{equation}

The fifth component centers on compliance with the limitations of the workspace of the MSP. It gets activated if a limit is exceeded. For simplification this limit is here defined as a maximal deviation of the pilot $x_\text{max}^\text{m} = 1.2$ m from the start position in a Cartesian coordinate system. A definition in a joint space system that accounts for the positional and dynamic limitations of the joints would further increase the efficient utilization of the workspace. Therefore, this should be considered in further investigations.
If this limit is exceeded, a high penalty is assigned:

\begin{equation}
	r_{ws,x} = \text{w}_{ws,x}
	\begin{cases}
		0 & \text{if } |x_t^\text{m}| < x_\text{max}^\text{m} \\
		1 & \text{else}.
	\end{cases}
\end{equation}

The same concept is implemented to achieve a rotational limit. In general, the use of a robotic-based MSP allows full rotation of the driver's cab \cite{2014Bellmann_BOOK}. However, this leads to an additional complication in the interpretability of the tilt coordination and is only advantageous in very specific application cases, such as an overhead flight. Therefore, in this work, the maximum absolute rotation is limited to $\varphi_\text{max}^\text{m} = \frac{\pi}{2}$:

\begin{equation}
	r_{ws,\varphi} = \text{w}_{ws,\varphi}
	\begin{cases}
		0 & \text{if } |\varphi_t^\text{m}| < \varphi_\text{max}^\text{m} \\
		1 & \text{else}.
	\end{cases}
\end{equation}

Additionally, when one of the both limitations is reached, the training episode is terminated and the robot is reset to the starting point $x_t^\text{m}=0.0$ m and $\varphi_t^\text{m}=0.0$. The values of the coefficients are determined by expert knowledge to $\text{w}_f = 0.33, \text{w}_\omega = 18.91, \text{w}_F = 0.67, \text{w}_x = 0.1, \text{w}_{ws,x} = 10,000, \text{w}_{ws,\varphi} = 10,000$. At each time step the total numeric reward $R_t$ and the state vector $S_t$ are getting transferred from the environment to the agent.

%% file: 03_Results_and_Discussion.tex
\section{Results and Discussion}
The framework is realized in python, using the library Stable Baselines3 \cite{2021Raffin_ARTC}, which is a PyTorch-based implementation of different RL algorithms. Besides the interface and the environment itself, the training process is specified here by the definition of various hyperparameters. In the following section, these points and the subsequent validation of the results will be further discussed.

\subsection{Training Configuration}

\begin{figure}
	\includegraphics[width=\columnwidth]{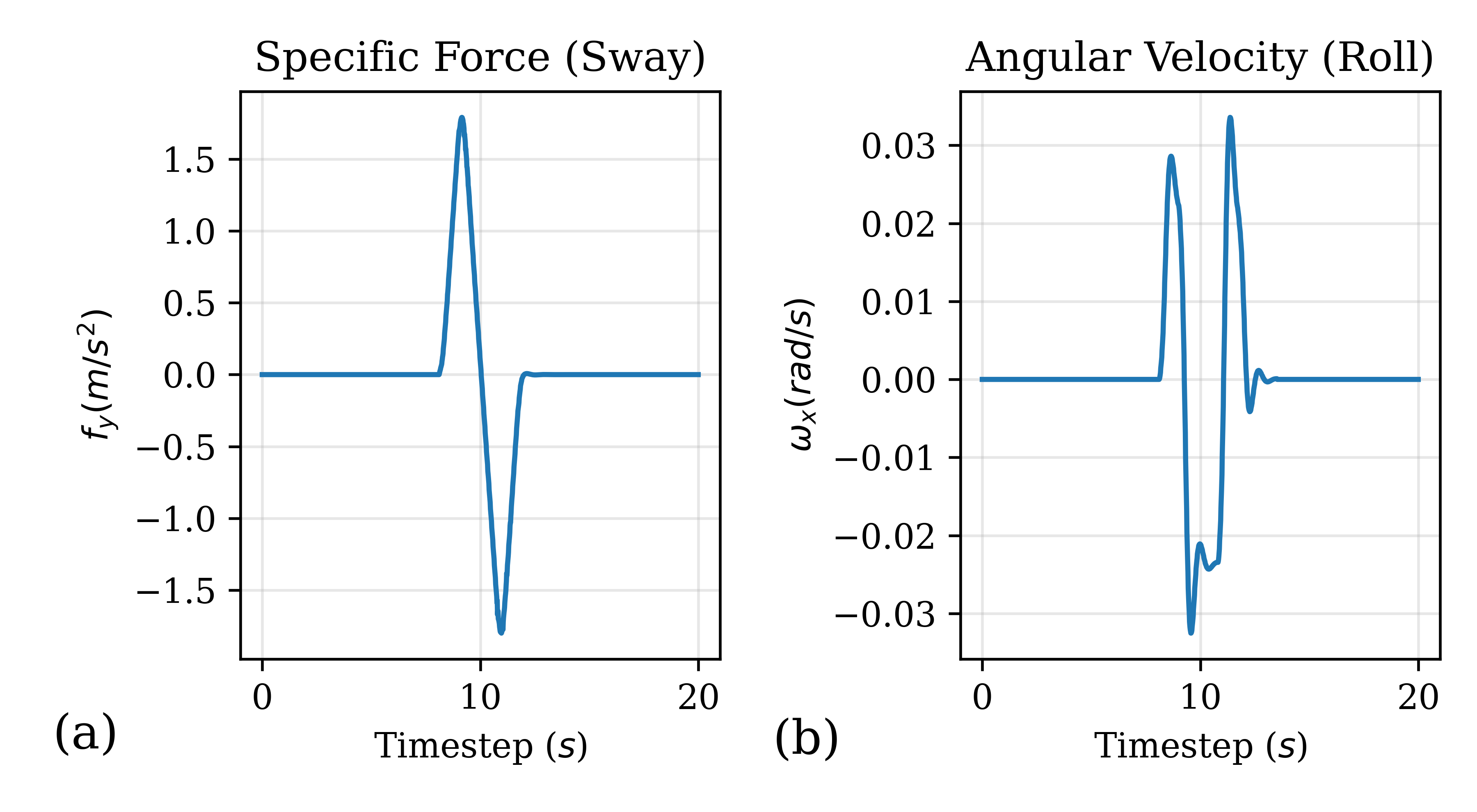}
	\caption{Plot of the lateral specific force $f_y$ (a) and the angular velocity $\omega_x$(b) around the roll axis acting on the driver while performing a lateral lane change to the left with a displacement of $2.84$~m and an initial driving speed $v_x$ of $8$~m/s. The therefore resulting maximal and minimal specific forces $f_{y,\text{min/max}}$ are $-1.80$~m/$\text{s}^2$ respectively $1.79$~m/$\text{s}^2$.}
	\label{f_w_single_lane_change_plot}
\end{figure}

Within the environment, the vehicle simulation specifies the target trajectories driven or flown by the vehicle. At each time step $t$ the translational specific force $f_{t+1}^\text{v}$ and the rotational angular velocity $\omega_{t+1}^\text{v}$ acting on the driver in the vehicle simulation are transferred as parts of the state to the agent. This, together with the rest of the state, provides the agent with information about the current kinematic state of the motion simulator and which signals should act on the pilot in the next time step in order to achieve a realistic immersion. Therefore, by specifying the target trajectories, the vehicle simulation determines the specific training data to which the agent learns to adapt its control strategy. In order to achieve a complete closed-loop MCA, this training data must cover a sufficient operational domain of potentially possible motions. In this work, this domain includes lateral movements of a vehicle. In a real-world scenario this corresponds to sudden evasive maneuvers or lane changes. The signal is generated by using a dynamics model of the DLR ROboMObil~\cite{2011Brembeck} created in Modelica, a specialized language for physical systems. One trajectory is recorded which is then artificially extended by a stochastic variation of the amplitude, the direction and the start time of the maneuver. Fig.~\ref{f_w_single_lane_change_plot} shows the plot of the resulting specific force $f_y$ and angular velocity $\omega_x$ of one example trajectory applied on the real car driver. The data is then integrated into the environment in the form of a lookup table.

The model of the MSP is also created in Modelica and implemented directly in the environment as a Functional Mock-up Unit (FMU), a software container being defined by the Functional Mock-up Interface standard~\cite{2021Junghanns}. Reducing the approach to sway and roll and restricting them in Cartesian space to configurations realizable by the MSP, allows it to be sufficient to only consider the position and orientation of the end effector during training. Both can be determined by Euler integration of the actions of the respective degree. For further use of the trajectory, the joint positions of the robot at each time step are calculated by an analytical solution of the inverse kinematics. However, their dynamics are not constrained in the work performed here, which would require the use of a subsequent controller to be able to follow the trajectories on a physical MSP. In order to exploit the full complexity of the workspace of a robotic MSP, a direct consideration and constraint of the joint angles already during training is necessary. This extension of the concept is therefore recommended for future work.

\begin{figure}
	\includegraphics[width=\columnwidth]{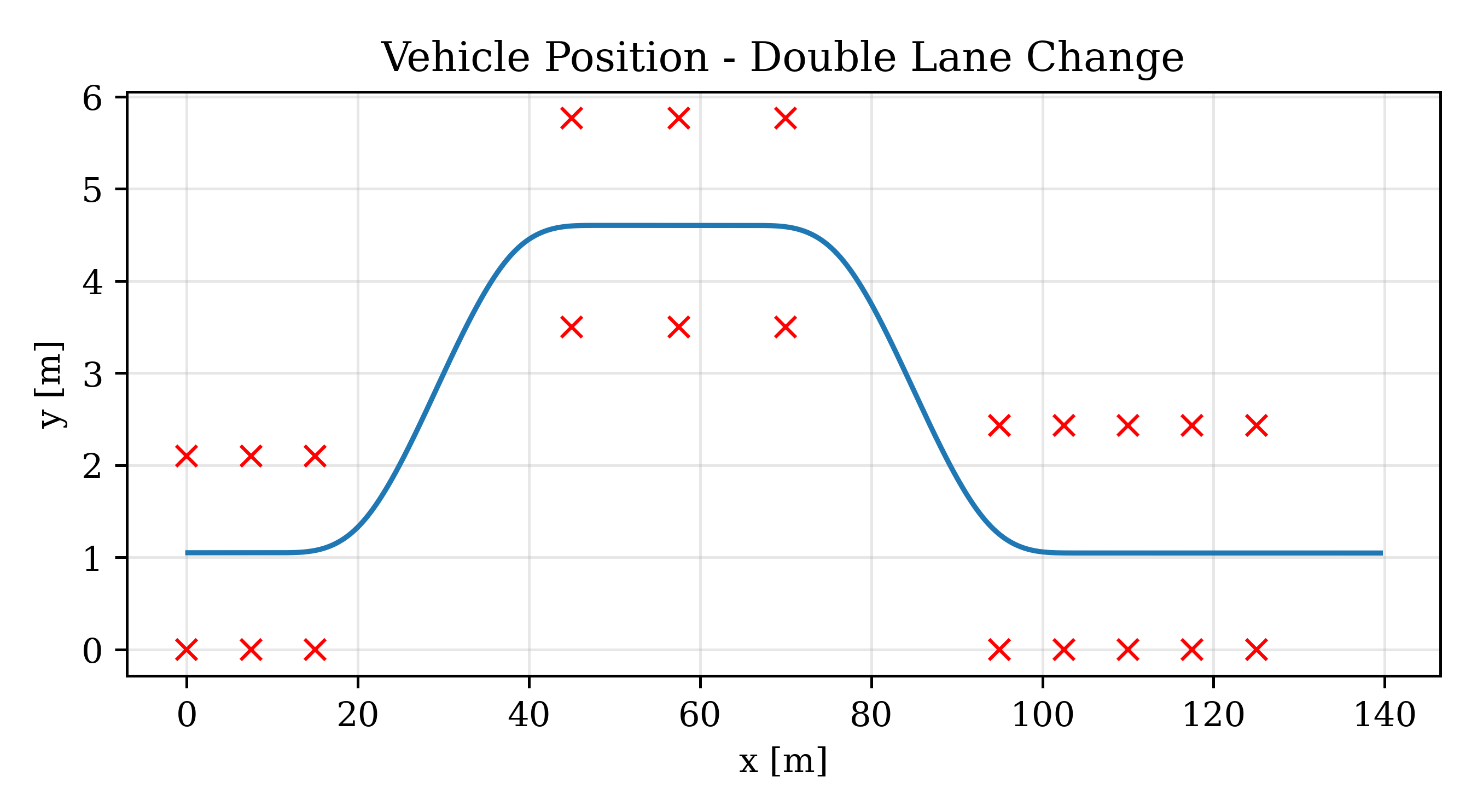}
	\caption{Test track and trajectory of a double lane change according to ISO-3888-1:2018 performed with a passenger car ($1.68$ m width) and a driving velocity of $10$ m/s. The red crosses represent traffic cones that mark out the test track and the blue line shows the driving trajectory.}
	\label{r_double_lane_change_plot}
\end{figure}

\begin{figure}
	\includegraphics[width=\columnwidth]{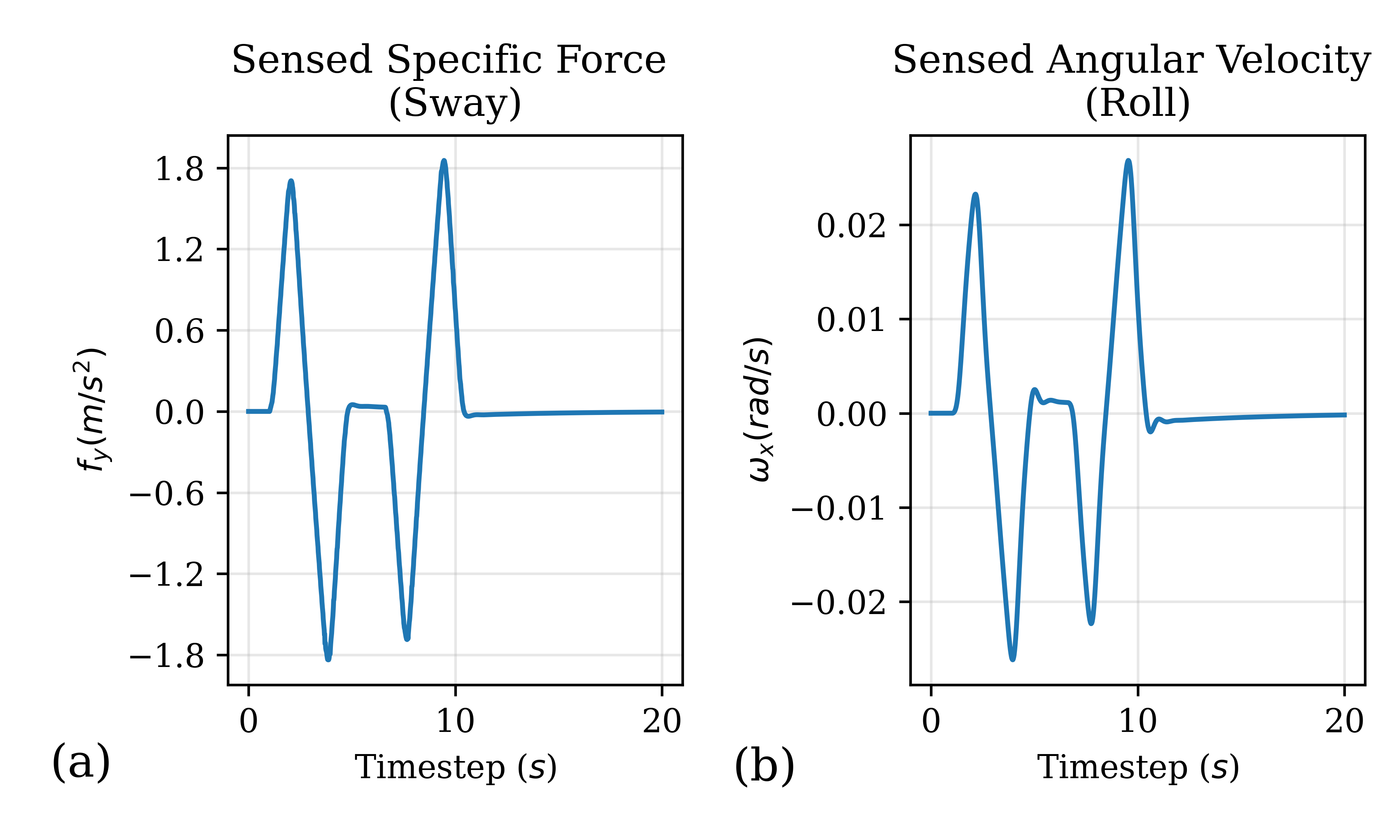}
	\caption{Plot of the sensed lateral specific force $f_y$ (a) and the sensed angular velocity $\omega_x$(b) around the roll axis acting on the driver while performing a double lane change according to ISO 3888-1:2018 with the DLR ROboMObil.}
	\label{f_w_double_lane_change_sensed_plot}
\end{figure}

The training process is then structured episodically, with each episode consisting of 5 sections, each of which includes one lane change. This setup ensures that the MSP is not returned to the initial position after each lane change, but starting the next section in an arbitrary state. Thereby a more stable training process takes place and the risk of overfitting gets reduced. Each section consists of $20.004$ s which corresponds to $1{,}667$ time steps with a step size of $12$ ms. The step size is set to match the step size of the MSP. The whole training process consists of $57$ million time steps, which corresponds to a simulated time of about $684{,}000$ s. The computation time of the training is about 7 hours on a core of an Intel Xeon Gold 6136 with 3.0 GHz clock speed. During the training process, the two ANNs are trained. Both networks are multilayer perceptrons with two hidden layers of 64 nodes each and ReLU activation functions. The used hyperparameters correspond to the default settings \cite{2021Raffin_ARTC} for the most part, with the following key changes. Since the PPO algorithm used is an on-policy algorithm that obtains its training data only from the current updated policy, a sufficiently large batch size must be selected for this data set. Therefore, the batch size chosen here is $2^{18}$ samples. The data set selected from the batch to update the neural network parameters, called mini-batch size in the implementation, is set to $2^{11}$ samples. To obtain a reasonable foresight of the agent, the discount factor $\gamma$ is set to $0.999$. The weighting coefficients of the reward function are determined in an empirical optimization process based on the best expert knowledge.

\subsection{Validation of the Results}

To perform a validation of the trained RL-based MCA, its performance is compared with the performance of a state of the art algorithm for a defined test case. The vehicle trajectory selected for this test case is a standardized double lane change according to ISO-3888-1:2018, as shown in Fig.~\ref{r_double_lane_change_plot}. The specific force $f_y$ and angular velocity $\omega_x$ sensed by the vehicle driver during this maneuver driven with the DLR ROboMObil are shown in Fig.~\ref{f_w_double_lane_change_sensed_plot}. This maneuver was not part of the data used during training and therefore demonstrates the ability of applying the MCA to unknown data within the known operational domain of trajectories.

\begin{figure*}
	\includegraphics[width=\textwidth]{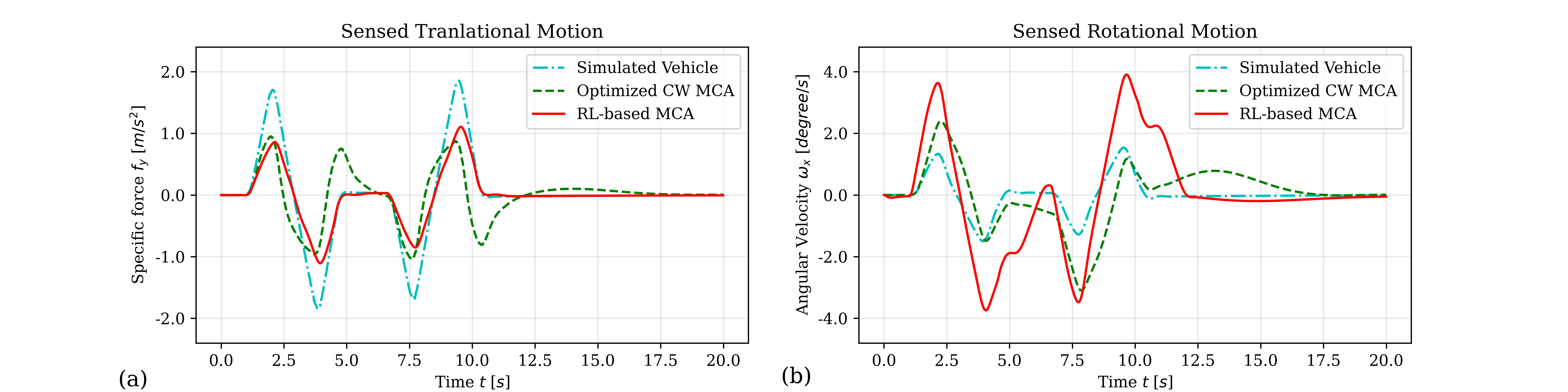}
	\caption{Comparison of the sensed translational (a) and rotational (b) motion for the optimized filter-based CW and the RL-based MCA.}
	\label{Perceived_signals}
\end{figure*}

\begin{figure*}
	\includegraphics[width=\textwidth]{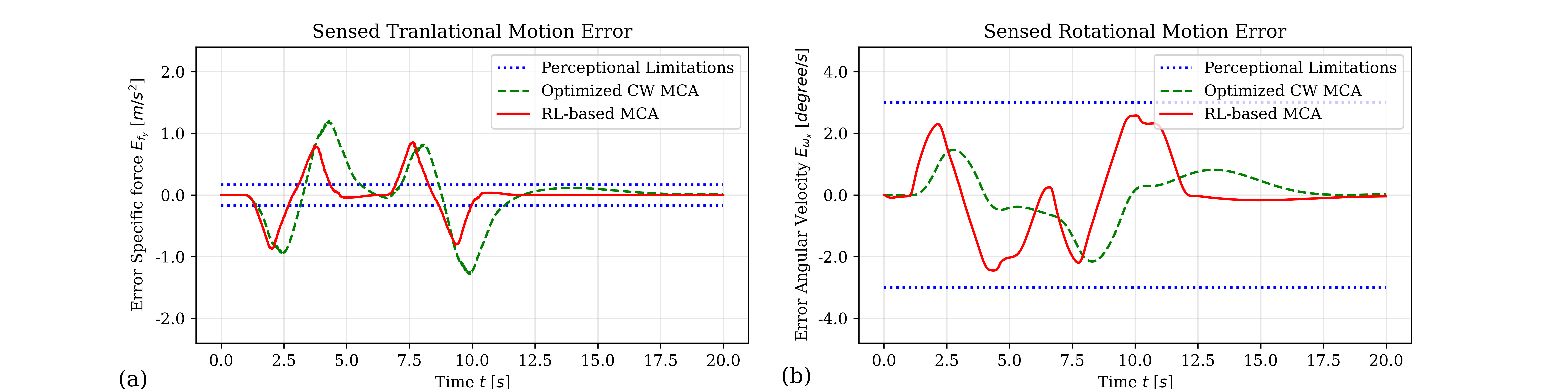}
	\caption{Comparison of the sensed translational (a) and rotational (b) motion error for the optimized filter-based CW and the RL-based MCA.}
	\label{Perceived_errors}
\end{figure*}

The chosen comparison algorithm is a specifically optimized filter-based CW. Therefore, the CW is used with optimized WF parameters for only this specific recorded driving scenario. This provides the best possible filter-based output for the available workspace to obtain the optimal regenerated motion to be compared with the RL-based algorithm. A Simplicial Homology Global Optimization (SHGO) algorithm \cite{2018Endres} with a Sequential Least Squares Programming (SLSQP) method \cite{Kraft1988} for local refinement is used to solve the constraint optimization problem, as proposed by Scheidel et al. in~\cite{2022Scheidel}. The variables of the optimization are the parameters of the filters $p$. The optimization problem with the cost function $J_\text{Opt}(p)$ and constraints is formulated as follows:

\begin{equation}
	\underset{p}{\min} \ J_\text{Opt}(p) = 
	\text{w}_\text{1} \hat{e}_f(p) + 
	\text{w}_\text{2} \hat{e}_\omega(p),
\end{equation}

\begin{equation}
	\text{s.t.} \ x_\text{min} < x^r(p) < x_\text{max},
	\label{constrains}
\end{equation}

where $\hat{e}_f (p) = (\hat{f}_\text{sens}^\text{v}(p) - \hat{f}_\text{sens}^\text{r})^2$ and $\hat{e}_\omega (p) = (\hat{\omega}_\text{sens}^\text{v}(p) - \hat{\omega}_\text{sens}^\text{r})^2$ are the squared errors in perception  and the weighting factors $\text{w}_\text{1} = 1.0$ and $\text{w}_\text{2} = 20.0$ are used to balance the optimization objectives, chosen according to expert knowledge. Here $\hat{f}_\text{sens}^\text{v}$, $\hat{f}_\text{sens}^\text{r}$, $\hat{\omega}_\text{sens}^\text{v}$ and $\hat{\omega}_\text{sens}^\text{r}$ are the sensed translational and rotational motions of the driver of the vehicle and the simulator user. The calculation of these signals is based on the mathematical model \cite{2005Telban_ARTC} of the vestibular system as follows

\begin{equation}
	\frac{\hat{f}_\text{sens}}{f} = \text{TF}_\text{oto} = \frac{K_{\text{oto}}(\tau_1 s+1)}{(\tau_2 s + 1) (\tau_3 s +1)},
\end{equation}

\begin{equation}
	\frac{\hat{\omega}_\text{sens}}{\omega} = \text{TF}_\text{scc} = \frac{\tau_4 \tau_5 s}{(\tau_4 s + 1) (\tau_5 s + 1)},
\end{equation}

where $\text{TF}_\text{oto}$ is the transfer function of the otoliths \cite{2016Asadi_ARTCa} and $\text{TF}_\text{scc}$ is the transfer function of the semicircular canals \cite{2017Asadi_ARTC}. The coefficients of this functions are chosen according to \cite{2005Telban_ARTC} as $K_{\text{oto}} = 0.4$, $\tau_1 = 5$ s, $\tau_2 = 0.016$ s, $\tau_3 = 10$ s, $\tau_4 = 5.73$ s and $\tau_5 = 80$ s in lateral direction. 

The cost function thus contains considerably less components than the RL training reward function given in Eq.~\ref{eq:rl_reward}. The cost function of CW optimization thus concentrates only on the two main objectives. The other components are not needed because they are already considered in the CW or in the optimization algorithm used. For example, no additional incentive is needed for washing out the signal, as this is permanently the case due to the application of the CW's high-pass filters. Compared to usual cost function in other optimizations of MCAs~\cite{2016Asadi_ARTC, 2020Qazani_ARTCa, 2022Asadi_ARTC, 2022Qazani} the use of workspace is not minimized. This is usually done to favor economic utilization of the workspace. However, since the goal of the optimization performed here is not the development of an optimal general-purpose MCA but the creation of an optimal filter-based comparison algorithm, its consideration here is not desirable However, it is necessary to take into account the limitations of the existing workspace. This is done by implementing the constrains in Eq.~\ref{constrains}. These can be considered directly by SHGO. For further refinement by SLSQP they are internally approximated by a smoothed differentiable equation. This ensures that only realizable solutions are taken into account. Furthermore, as usual for CW filters, the angular velocity resulting from tilt coordination, is limited to the perception threshold of $0.0524$ rad/s according to \cite{1985Reid_ARTC}.

\begin{figure}
	\includegraphics[width=\columnwidth]{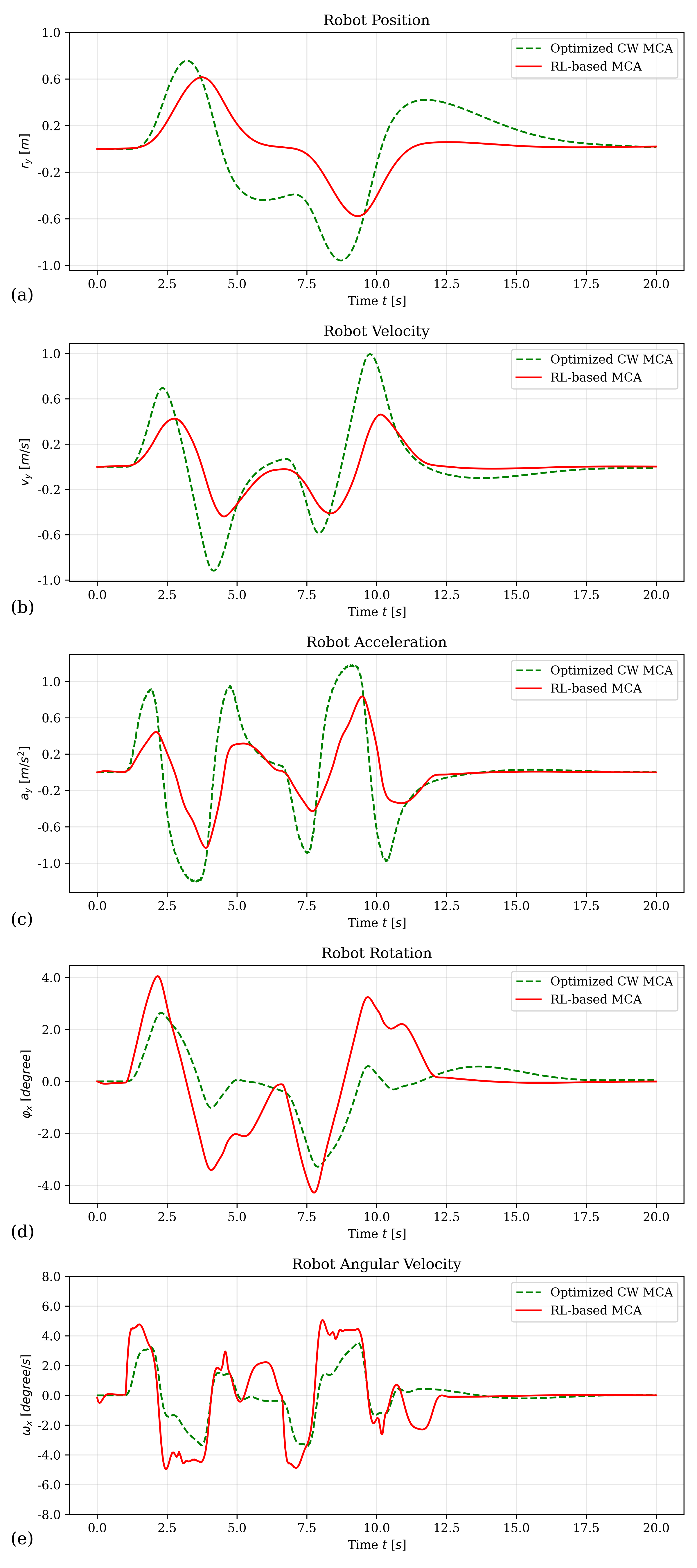}
	\caption{Performed trajectories of the end effector of the robotic motion simulator for the optimized CW MCA and the RL-based MCA. The translational position (a), velocity (b) and acceleration (c) and the rotation (d) and angular velocity (e) are shown.}
	\label{robot_comparison}
\end{figure}

Fig.~\ref{Perceived_signals} shows the different perceived signals of the pilot using these two algorithms in comparison with the motion sensation of the real vehicle driver (reference signal). Both algorithms achieve comparable magnitudes for the lateral specific force. However, the shape of the perceived signal produced by the RL-based MCA follows the shape of the reference signal considerably better compared to the optimized CW. Specifically, the false cues (at around $t=4.5$ s and $t=10.5$ s) that are generated by the optimized CW filter are decreased remarkably using the RL-based MCA. This is of particular significance as these can lead to the occurrence of motion sickness. Additionally, Fig. \ref{Perceived_errors} shows the errors of the sensed signals of both algorithms and the perceptional limitations \cite{1985Reid_ARTC}. Here it can be seen that the resulting error of the sensed angular velocity for both algorithms stays within those defined perceptual limitations. The RL-based MCA approaches the threshold limitation more closely and thus makes better use of the available margin to perform the tilt coordination compared with the optimized CW filter. In addition, the use of the optimized CW filter leads to a significant delay of the signal.

\begin{table}\tiny
	\centering
	\rowcolors{2}{gray!10!}{white}
	\caption{Comparison of the metrics of both MCAs. The perceptional thresholds are $0.17 \frac{\text{m}}{\text{s}^2}$ translational in lateral and $3.0 \frac{\text{degree}}{\text{s}}$ rotational in roll direction \cite{1985Reid_ARTC}}
	\label{metrics}
	\begin{tabular}{>{\bfseries} r r r}
		& \textbf{RL-based MCA} & \textbf{Optimized CW Filter} \\
		\hline
		Sensed translational RMS error & 0.289 m/$\text{s}^\text{2}$ & 0.469 m/$\text{s}^\text{2}$ \\
		Sensed rotational RMS error & 1.226 degree/s & 0.776 degree/s\\
		Translational correlation coefficient & 0.990 & 0.715 \\
		Rotational correlation coefficient & 0.860 & 0.623 \\
	\end{tabular}
\end{table}

In Table~\ref{metrics} the performance metrics of the RL-based MCA and the optimized CW are presented. These are the RMS errors and the correlation coefficients of the signals sensed using the two algorithms compared to the sensed reference signals. The consideration of these metrics is based on the work of Asadi et al. \cite{2016Asadi_ARTC}. They proposed the use of Pearson's correlation coefficient to evaluate the shape following in addition to the evaluation of the RMS error. The implementation of the RL-based MCA leads to an amelioration of the value of the correlation coefficient for both rotational and translational channel directions in comparison to the use of the optimized CW MCA. The RMS error of the sensed specif force between the real vehicle driver and pilot can be reduced for the translational motions. This is also due to the increase in the usage of the tilt coordination, which in turn is reflected in the rise of the rotational RMS error. However, this has no negative impact on the quality of the simulation, since the error, as shown for the whole trajectory in Fig.~\ref{Perceived_errors}, is below the perception threshold.

Fig.~\ref{robot_comparison} shows the movements of the MSP when performing the double lane change using the two different algorithms. The use of the available translational workspace is lower for the implementation of the RL-based MCA. The maximum displacement with respect to the center can be reduced from $0.96$ m for optimized CW filter to $0.62$ m for RL-based MCA. The same reduction can be seen for the occurring velocities and accelerations. The examination of the rotational motion using RL-based MCA shows an increase, which again can be assigned to the tilt coordination usage.
Overall, it can be shown that within the trained dataset, an open loop ANN-based MCA with the capability of running in real-time can be obtained using RL, which achieves an improvement in motion simulation compared to an optimized washout filter-based algorithm.

%% file: 04_Conclusion.tex
\section{Conclusion}
The quality of immersion of a motion simulation using MSPs depends severely on the implemented MCA. Application of this algorithm is necessary to optimize the use of the limited available workspace to represent real vehicle movements in a simulated environment.
In this paper, a new unconventional approach is proposed using deep RL to find the optimal motions of the MSP. This can lead to an ANN-based MCA that can perform real-time positioning of a motion simulator within the set of trained inputs. Thereby, problems of existing MCAs, such as the real-time capability of MPC-based MCAs and the relatively poor motion illusion of filter-based algorithms, are solved simultaneously.
To achieve this, an MDP was formulated that allows the RL agent to communicate with the designed environment through a defined interface. A PPO algorithm, which uses the transmitted feedback in a training process, is applied to gradually gain an understanding of the environment's behavior. The learned control strategy stored in an ANN corresponds to an MCA.

To demonstrate the functionality of RL-based MCA, a specific use case is created where stochasticity is used to add variance to the input trajectories.
These are various lateral movements that resemble a lane change or an evasive maneuver. The time, direction and amplitude of these movements are stochastically selected. The resulting lateral translational acceleration and angular velocity are then used as reference signals in the training process. The trained RL-based MCA is thus capable of performing an optimized movement of the motion simulator for any lateral motion within the defined set. For validation, the standardized maneuver of a double lane change is used. A CW filter optimized specifically for this trajectory is used as a benchmark algorithm. Evaluation of the perceived acceleration and angular velocity shows that RL-based MCA achieves improvement in motion while reducing the linear workspace usage of the MSP. The correlation coefficients of both translational and rotational motion directions increase, the translational RMS error decreases and the error of the angular velocity remains within the defined perception threshold. Furthermore, the false cues that occur when using the optimized CW could be successfully eliminated. Hence, both the feasibility of applying RL to MCAs and the resulting improvement in motion simulation can be demonstrated.

Further work can investigate an extension of the training data to obtain a more versatile MCA. Furthermore, a consideration of the vestibular system in the training process could improve the performance and a change of the limitation of the workspace from Cartesian to joint angle space could be investigated. In addition to extending the functionality of the MCA, future studies should also look more closely at the deep RL methodology used. In addition to comparing different training algorithms, the choice of training hyperparameters may also be investigated.